\documentclass{article}

\newcommand\arxiv{}

\usepackage{ijcai21}

\usepackage{times}
\usepackage{soul}
\usepackage{url}
\usepackage[hidelinks]{hyperref}
\usepackage[utf8]{inputenc}
\usepackage[small]{caption}
\usepackage{graphicx}
\usepackage{amsmath}
\usepackage{amsthm}
\usepackage{booktabs}
\usepackage{algorithm}
\usepackage{algorithmic}
\usepackage{latexsym}

\usepackage{amssymb}
\usepackage{bbm}

\usepackage{tikz}
\usetikzlibrary{automata}

\usepackage{xspace}
\usepackage{xcolor}
\usepackage{booktabs}
\usepackage{colortbl}
\usepackage{tabularx}

\usepackage{bm}
\renewcommand{\vec}[1]{\bm{#1}}

\newtheorem{theorem}{Theorem}
\newtheorem{lemma}{Lemma}
\newtheorem{corollary}{Corollary}

\theoremstyle{definition}

\newtheorem{example}{Example}
\newcommand{\qee}{\hfill$\triangle$}
\theoremstyle{remark}

\newcommand{\Qset}{\mathbb{Q}}
\newcommand{\Nset}{\mathbb{N}}
\newcommand{\Rset}{\mathbb{R}}

\newcommand{\vecl}[1]{\vec{\mathit{#1}}}
\newcommand{\dist}{\mathit{Dist}}

\newcommand{\X}{{\ensuremath{\mathbf{X}}}}
\newcommand{\F}{{\ensuremath{\mathbf{F}}}}
\newcommand{\G}{{\ensuremath{\mathbf{G}}}}
\newcommand{\U}{{\ensuremath{\mathbf{U}}}}

\newcommand{\inits}{\hat s}
\newcommand{\act}[1]{\mathit{Act}(#1)}
\newcommand{\pat}{\rho}
\newcommand{\Pat}{\mathsf{Runs}}
\newcommand{\fpat}{w}
\newcommand{\mem}{\mathsf{M}}
\newcommand{\Cone}{\mathsf{Cone}}
\newcommand{\calF}{\mathcal{F}}
\newcommand{\mec}{\mathsf{MEC}}

\newcommand{\pr}{\mathbb P}	
\renewcommand{\Pr}[3]{\pr^{#1}\hspace{-0.16em}\left[{#3}\right]}   
\newcommand{\expected}{\mathbb E}
\newcommand{\Ex}[3]{\expected^{#1}_{#2}\hspace{-0.16em}\left[{#3}\right]}   

\newcommand{\reward}{{r}}

\newcommand{\lrLim}[1]{\mathrm{lr}(#1)}  
\newcommand{\lrSf}[1]{\mathrm{lr}_{\mathrm{sup}}(#1)}  
\newcommand{\lrIf}[1]{\mathrm{lr}_{\mathrm{inf}}(#1)}  

\newcommand{\lrs}{\mathfrak {LR}}
\newcommand{\sss}{\mathcal S}
\newcommand{\rs}{R}
\newcommand{\ps}{\theta}
\newcommand{\mdp}{\mathcal G}
\newcommand{\mc}{\mathcal M}
\newcommand{\auto}{\mathcal{LDBA}(\varphi)}
\newcommand{\rmax}{R_{\max}}
\newcommand{\tran}[1]{\stackrel{#1}{\to}}

\newcommand{\amec}{\mathsf{AMEC}}
\newcommand{\acc}{\mathit{AccA}}
\newcommand{\lang}{\mathsf L}
\newcommand{\pair}[2]{\langle#1,#2\rangle}

\newcommand{\kda}{\mathbf{1}_a} 

\newcommand{\vfreq}{\vecl{freq}}
\newcommand{\freq}{\mathrm{freq}}

\usepackage{microtype}

\ifdefined\arxiv
\fi

\newcommand{\spacefu}{\vspace*{-1.2em}}
\newcommand{\spacefl}{\vspace*{-0.8em}}

\newcommand{\paraa}[1]{\smallskip\noindent{\bf #1}}
\newcommand{\para}[1]{\paragraph{#1.}}

\title{LTL-Constrained Steady-State Policy Synthesis}
\author{
    Jan K\v ret\'insk\'y
    \affiliations
    Technical University of Munich	
    \emails
    jan.kretinsky@tum.de
}

\begin{document}

\maketitle

\begin{abstract}
  Decision-making policies for agents are often synthesized with the constraint that a  formal specification of behaviour is satisfied. 
  Here we focus on infinite-horizon properties.
  On the one hand,  Linear Temporal Logic (LTL) is a popular example of a formalism for qualitative specifications.
  On the other hand, Steady-State Policy Synthesis (SSPS) has recently received considerable attention as it provides a more quantitative and more behavioural perspective on specifications, in terms of the frequency with which states are visited.
  Finally, rewards provide a classic framework for quantitative properties.
  In this paper, we study Markov decision processes (MDP) with the specification combining all these three types.
  The derived policy maximizes the reward among all policies ensuring the LTL specification with the given probability and adhering to the steady-state constraints.
  To this end, we provide a unified solution reducing the multi-type specification to a multi-dimensional long-run average reward.
  This is enabled by Limit-Deterministic B\"uchi Automata (LDBA), recently studied in the context of LTL model checking on MDP, and allows for an elegant solution through a simple linear programme.
  The algorithm also extends to the general $\omega$-regular properties and runs in time polynomial in the sizes of the MDP as well as the LDBA.
\end{abstract}

\section{Introduction}

\paraa{Markov decision processes} (MDP), e.g.~\cite{puterman}, are a basic model for agents operating in uncertain environments since it features both non-determinism and stochasticity.
An important class of problems is to automatically synthesize a policy that resolves the non-determinism in such a way that a given formal specification of some type is satisfied on the induced Markov chain.
Here we focus on the classic formalisms capable of specifying properties over \emph{infinite} horizon.

\para{Infinite-horizon properties}
On the one hand, \emph{Linear Temporal Logic} (LTL) \cite{pnueli77} is a popular formalism for qualitative specifications, capable of expressing complex temporal relationships, abstracting from the concrete quantitative timing, e.g., \emph{after every request there is a grant} (not saying when exactly).
It has found applications in verification of programs \cite{BK08} as well as high-level robot control, e.g.\ \cite{DBLP:journals/trob/Kress-GazitFP09}, or preference-based planning in PDDL \cite{DBLP:conf/aips/BentonCC12} to name a few.

On the other hand, \emph{Steady-State Control} (SSC, a.k.a.\ Steady-State Policy Synthesis) \cite{akshay13} constrains the frequency with which states are visited, providing a more quantitative and more behavioural perspective (in terms of states of the system, as opposed to logic-based or reward-based specifications).
Recently, it has started receiving more attention also in AI planning \cite{ijcai19,ijcai20}, improving the theoretical complexity and its applicability to a wider class of MDP (although still being quite restrictive on the class of policies, see below).

Finally, rewards provide a classic framework for quantitative properties.
In the setting of infinite horizon, a key role is played by the \emph{long-run average reward} (LRA), e.g.~\cite{puterman}, which constrains the reward gained on average per step and,
over the decades, it has found numerous applications \cite{feinberg2012handbook}.

\para{Our contribution}
In this paper, we study Markov decision processes (MDP) with the specifications \emph{combining all these three types}, yielding a more balanced and holistic perspective.
We synthesize a policy maximizing the LRA reward among all policies ensuring the LTL specification (with the given probability) and adhering to the steady-state constraints. 
To this end, we provide a \emph{unified solution} conceptually reducing the problem with the heterogeneous specification combining three types of properties to a problem with a single-type specification, namely the multi-dimensional LRA reward.
This in turn can be solved using a single simple linear programme, which is easy to understand and shorter than previous solutions to the special cases considered in the literature.

Not only does the unified approach allow us to use the powerful results from literature on rewards, but it also allows for easy extensions and modifications such as considering various classes of policies (depending on the memory available or further requirements on the induced chain) or variants of the objectives (e.g.~constraints on frequency of satisfying recurrent goals, 
 multiple rewards, trade-offs between objectives), which we also describe.

Our reduction is particularly elegant due to the use of  Limit-Deterministic B\"uchi Automata (LDBA), recently studied in the context of LTL model checking on MDP.
Moreover, the solution extends trivially from LTL to the general $\omega$-regular properties (given as automata) and runs in time polynomial in the sizes of the MDP as well as the LDBA.

In summary, our contribution is as follows:
\begin{itemize}
	\item We introduce the heterogeneous LTL-SSC-LRA specification for maximizing the reward under the LTL and steady-state constraints.
	\item We provide a unified solution framework via recent results on LDBA automata and on multi-dimensional LRA optimization.
	\item The resulting solution is generic, as documented on a number of extensions and variants of the problem we discuss, as well as simple, generating a linear programme with a structure close to classic reward optimization.
\end{itemize}

\subsection{Related Work}

The \emph{steady-state control} was introduced in \cite{akshay13}, treating the case of recurrent MDP and showing the problem is in PSPACE by quadratic programming.
It is combined with LRA reward maximization, giving rise to \emph{steady-state policy synthesis}, in \cite{ijcai19}.
A polynomial-time solution is provided via linear programming, even for general (multi-chain) MDP, but restricting to stationary policies inducing recurrent Markov chains.
In \cite{ijcai20}, general (multi-chain) MDP and a wider class of policies are considered.
However, the ``EP'' policies prohibit playing in some non-bottom end components,
which makes stationary policies optimal within this class, but is very restrictive in contrast to general strategies.
In this work, we add also the LTL specification and we have neither restriction on the MDP nor on the class of strategies considered and chains induced.
Our linear programme is simpler and extends the correspondence between solutions and the policies with their steady-state distributions to the general ones (with no recurrence assumed).
On the other hand, we also treat the restricted settings of finite-memory, recurrence, and further modifications to show the versatility of our approach.

The main idea of our approach is an underlying conceptual reduction to \emph{multi-dimensional LRA rewards} allowing for a simple linear programme.
The classic linear-programming solution for a single LRA reward \cite{puterman} has been extended to various settings such as multi-dimensional reachability and $\omega$-regular properties \cite{EKVY08} or multi-dimensional LRA rewards \cite{lmcs14,lmcs17}.
Some of these results \cite{lmcs14} are also implemented \cite{multigain} on top of the PRISM model checker \cite{prism}.
Rewards have also been combined with LTL on finite paths in \cite{DBLP:conf/aaai/BrafmanGP18,DBLP:conf/aips/GiacomoIFP19}.
Our results strongly rely particularly on \cite{lmcs17};
however, that work treats neither SSC nor LTL.

The other element simplifying our work is the \emph{limit-deterministic B\"uchi automaton} \cite{CY95}.
It has been shown to be usable for LTL model checking on MDP under some conditions \cite{concur15,cav16}, these conditions are satisfied by an efficient translation from LTL \cite{cav16}.
Tools for the translation \cite{rabinizer} as well as the model-checking \cite{mochiba} are also available.

\section{Preliminaries}

\subsection{Basic Definitions}

We use  $\Nset,\Qset,\Rset$ to denote the sets of positive integers, rational and real numbers, respectively. 
The Kronecker function $\mathbf 1_x(y)$ yields 1 if $x=y$ and 0 otherwise.
The set of all distributions over a countable set $X$ is denoted by $\dist(X)$.
A distribution $d\in\dist(X)$ is Dirac on $x\in X$ if $d=\mathbf1_{x}$.

\para{Markov chains} 
A \emph{Markov chain} (MC) is a tuple \mbox{$\mc = (L,P,\mu,\nu)$} where $L$ is a countable set of locations, 
$P:L\to\dist(L)$ is a probabilistic transition function,
$\mu\in\dist(L)$ is the initial probability distribution, and
$\nu:L\to 2^{Ap}$ is a labelling function.

A \emph{run} in $\mc$ is an infinite sequence $\pat = \ell_1 \ell_2 \cdots$ of locations,
a \emph{path} in $\mc$ is a finite prefix of a run. 
Each path $\fpat$ in $\mc$ determines the set $\Cone(\fpat)$ consisting of all runs that start with $\fpat$. 
To $\mc$ we associate the probability space $(\Pat,\calF,\pr)$, 
where $\Pat$ is the set of all runs in $\mc$, $\calF$ is the $\sigma$-field generated by all $\Cone(\fpat)$,
and $\pr$ is the unique probability measure such that
$\pr(\Cone(\ell_1\cdots\ell_k)) = 
\mu(\ell_1) \cdot \prod_{i=1}^{k-1} P(\ell_i)(\ell_{i+1})$.

\para{Markov decision processes} 
A \emph{Markov decision process} (MDP) is a tuple $\mdp=(S,A,\mathit{Act},\delta,\inits,\nu)$ 
where $S$ is a finite set of states, $A$ is a finite set of actions, 
$\mathit{Act} : S\rightarrow 2^A\setminus \{\emptyset\}$ assigns to each state $s$ the set $\act{s}$ of actions enabled 
in $s$ so that $\{\act{s}\mid s\in S\}$ is a partitioning of $A$,
$\delta : A\rightarrow \dist(S)$ is a probabilistic 
transition function that given 
an action $a$ 
gives a probability distribution over the 
successor states, $\inits$ is the initial state, and $\nu:S\to2^{Ap}$  is a labelling function. 
Note that every action is enabled in exactly one state (w.l.o.g.\ by renaming).

A \emph{run} in $\mdp$ is an infinite alternating sequence of states
and actions $\pat=s_1 a_1 s_2 a_2\cdots$
such that for all $i \geq 1$, we have $a_i\in\act{s_i}$ and $\delta(a_i)(s_{i+1}) > 0$. 
A \emph{path} of length~$k$ in~$\mdp$ is a finite prefix
$\fpat = s_1 a_1\cdots a_{k-1} s_k$ of a run in~$\mdp$.

\para{Policies and plays} 
Intuitively, a policy (a.k.a.\ strategy) in an MDP is a ``recipe'' to choose actions.
A policy is formally defined as a function 
$\sigma : (SA)^*S \to \dist(A)$ that given a finite path~$\fpat$, representing 
the history of a play, gives a probability distribution over the 
actions enabled in the last state of $\fpat$.

A \emph{play} of $\mdp$ determined by 
a policy $\sigma$ is a Markov chain 
$\mdp^\sigma$ where the set of locations is the set of paths $S(AS)^*$, the valuation depends on the current state only $\nu(\fpat s)=\nu(s)$,
the initial distribution is Dirac on $\inits$, and
transition function $P$ is defined by
\( 
P(\fpat)(\fpat a s)\ =\ \sigma(\fpat)(a)\cdot \delta(a)(s)\,. 
\)
Hence, $\mdp^\sigma$ starts in the initial state of the MDP and actions are chosen according to $\sigma$ evaluated on the current history of the play, and the next state is chosen according to the transition function evaluated on the chosen action. 
The induced probability measure is denoted by $\pr^{\sigma}$ and ``almost surely'' (``a.s.'') or ``almost all runs'' refers to happening with probability 1 according to this measure.
The respective expected value of a random variable $F:\Pat\to\Rset$ is $\expected^\sigma[F]=\int_\Pat F\ d\,\pr^\sigma$.
For $t\in\Nset$, random variables $S_t,A_t$ return the $t$-th state and action on the run.

A policy is \emph{memoryless} if it only depends on the last state of the current history.
A policy is \emph{bounded-memory} (or \emph{finite-memory}) if it stores only finite information about the history; technically, it can written down using a finite-state machine, as detailed in Appendix~\ref{app:strat}.

\para{End components}
A set $T\cup B$ with $\emptyset\neq T\subseteq S$ and $B\subseteq \bigcup_{t\in T}\act{t}$
is an \emph{end component} of $\mdp$
if (1) for all $a\in B$, whenever $\delta(a)(s')>0$ then $s'\in T$;
and (2) for all $s,t\in T$ there is a path 
$\fpat = s_1 a_1\cdots a_{k-1} s_k$ such that $s_1 = s$, $s_k=t$, and all states
and actions that appear in $\fpat$ belong to $T$ and $B$, respectively.
An end component $T\cup B$ is a \emph{maximal end component (MEC)}
if it is maximal with respect to the subset ordering. 
Given an MDP, the set of MECs is denoted by $\mec$ and can be computed in polynomial time~\cite{CY95}.

\subsection{Specifications}\label{ssec:spec}

In order to define our problem, we first recall definitions of the considered classes of properties.

\para{LTL and $\omega$-regular properties}
\emph{Linear Temporal Logic} (LTL) \cite{pnueli77} over the finite set $Ap$ of atomic propositions is given by the syntax 
\[
\varphi::=p\mid\varphi\wedge\varphi\mid\neg\varphi\mid \X\varphi\mid\varphi\U\varphi\qquad p\in Ap
\]
and is interpreted over a run $\pat=s_1s_2\cdots$ of $\mc$ by 
\begin{tabular}{ll}
	$\pat\models p$ &if $p\in L(s_1)$\\
	$\pat\models \X \varphi$ &if $s_2s_3\cdots\models\varphi$\\
	$\pat\models \varphi\U\psi$ &if $\exists k:s_ks_{k+1}\cdots\models\psi,\forall j<k: s_js_{j+1}\cdots\models\varphi$
\end{tabular}
and the usual semantics of Boolean connectives.
The event that $\varphi$ is satisfied is denoted by $\models\varphi$.
Given an MDP $\mdp$ and a policy $\sigma$, $\pr^\sigma[\models\varphi]$ then denotes the probability that $\varphi$ is satisfied on $\mdp^\sigma$.

A \emph{B\"uchi Automaton} (BA) is a tuple  $\mathcal{B} = (\Sigma, Q, \Delta, q_0, F)$
where $\Sigma$ is a finite alphabet, $Q$ is a finite set of states, $\Delta \colon  Q \times \Sigma \rightarrow 2^Q$ is a transition function, writing $q\tran{\alpha}r$ for $r\in\Delta(q,\alpha)$, $q_0$ is the initial state, and $F \subseteq Q$ is the set of accepting states.
$\mathcal{B}$ \emph{accepts} the language  $\lang(\mathcal B)=\{\alpha_1\alpha_2\cdots\in\Sigma^\omega\mid\exists q_0\tran{\alpha_1}q_1\tran{\alpha_2}q_2\cdots:q_i\in F\text{ for infinitely many }i\}$ of infinite words that can pass through $F$ infinitely often.

A BA $\mathcal{B}$ is {\em limit-deterministic} (LDBA) \cite{cav16} if $Q$ can be partitioned 
$Q = Q_\mathcal{N} \uplus Q_\mathcal{D}$ so that
\begin{enumerate}
	\item $\Delta(q, \alpha) \subseteq Q_\mathcal{D}$ and $|\Delta(q, \alpha)| = 1$ for 
	$q \in  Q_\mathcal{D}$, $\alpha \in \Sigma$,
	\item $F \subseteq Q_\mathcal{D}$, and
	\item $|\Delta(q,\alpha)\cap Q_{\mathcal N}|=1$ for 
	$q \in  Q_\mathcal{N}$, $\alpha \in \Sigma$.
\end{enumerate}
Intuitively, a BA is LDBA if it can be split into a non-deterministic part without accepting transitions and a deterministic component, where it has to remain forever but may now accept.
The third constraint is less standard and not essential, but eases the argumentation.
Intuitively, it implies that the only choice that is made on a run is when and where to do the single non-deterministic ``jump'' from $Q_\mathcal{N}$ to $Q_\mathcal{D}$ \cite{cav16}.

Every LTL formula can be translated into an equivalent BA $BA(\varphi)$ in exponential time and into LDBA $LDBA(\varphi)$ in double exponential time \cite{cav16}, so that both automata are over the alphabet $2^{Ap}$ and accept exactly the language $\lang(\varphi)$ of words satisfying $\varphi$.\footnote{Technically, $\lang(\varphi)$ is the set such that for any run $\pat=s_1s_2\cdots$ with $\nu(s_i)=\alpha_i$, we have $\pat\models\varphi$ iff $\alpha_1\alpha_2\cdots\in\lang(\varphi)$.}

A set is an $\omega$-regular language if it can be written as $\lang(\mathcal B)$ for some BA (or equivalently LDBA) $\mathcal B$.
The standard algorithm for LTL model checking on MDP, i.e. to decide whether $\exists\sigma:\pr^\sigma[\models\varphi]\geq p$, is to translate it to the more general problem of model checking $\omega$-regular specification $\exists\sigma:\pr^\sigma[\nu(\pat)\in\lang(\mathcal A)]\geq p$ for some automaton $\mathcal A$ accepting $\lang(\varphi)$.
Typically, $\mathcal A$ is deterministic with a more complex acceptance condition \cite{BK08}, but also LDBA (with the simple B\"uchi condition of visiting $F$ infinitely often) can be used if they satisfy a certain condition, ensured by the recent translation from LTL \cite{cav16}.
Then the problem can be solved by combining the MDP $\mdp$ and the automaton $\mathcal B:=LDBA(\varphi)$ into the \emph{product} $\mdp\times \mathcal B$, where the automaton ``monitors'' the run of the MDP, and its subsequent analysis.
The product of $\mdp=(S,A,\mathit{Act},\delta,\inits,\nu)$ and $\mathcal{B} = (2^{Ap}, Q, \Delta, q_0, F)$ is the MDP $\mdp\times \mathcal B=(S\times Q, A', \mathit{Act}', \delta', \pair{\inits}{q_0}, \nu')$ where
\begin{itemize}
	\item $A'=A\times Q\times Q$ ensuring uniqueness of actions, the elements are written as $a_{q\to r}$, intuitively combining $a$ with $q\to r$;
	\item $\mathit{Act}'(\pair sq)= \{a_{q\to r} \mid a\in\act{s}, q\tran{\nu(s)}r\}$
	\item $\delta'(\pair sq,a_{q\to r})(\pair t{r'}) = \delta(s,a)(t)\cdot\mathbf 1_{r}(r')$
	\item $\nu'(\pair sq)=\nu(s)$
\end{itemize}
Note that indexing the actions preserves their uniqueness.
A state $\pair sq$ of the product is \emph{accepting} if $q\in F$.
Observe that a run of the original MDP satisfies $\varphi$ iff there is a choice of the jump such that the corresponding run of the product visits accepting states infinitely often.
A MEC which contains an accepting state is called an \emph{accepting MEC} and the set of accepting MECs is denoted by $\amec$.
Observe that once in $\amec$, any policy that uses all actions of the current MEC (and none that leave it) ensures visiting all its states hence also acceptance.
Consequently, $\max_\sigma\Pr{\sigma}{\mdp}{\models\varphi}=\max_\sigma\Pr{\sigma}{\mdp\times\auto}{\bigcup\amec\text{ is reached}}$ by \cite{cav16}.

\para{Steady-state constraints}
The classic way of constraining the steady state is to require that $\pi(s)\in[\ell,u]$ where $\pi(s)$ is the steady-state distribution applied to the state $s$ and $[\ell,u]\subseteq[0,1]$ is the constraining interval.
We slightly generalize the definition in two directions: 
Firstly, we can constrain (not necessarily disjoint) \emph{sets} of states, to match the ability of the logical specification. 
Secondly, we consider general cases where the steady-state distribution does not exist, to cater for general policies (in contrast to restricted classes of policies investigated in previous works), which may cause such behaviour.
Formally, a \emph{steady-state specification} (SSS) is a set $\sss$ of steady-state constraints of the form $(p,\ell,u)\in Ap\times[0,1]\times[0,1]$.
A Markov chain $\mc$ satisfies $\sss$, written $\mc\models\sss$ if for each $(p,\ell,u)\in\sss$ we have
\[\arraycolsep=1.4pt\begin{array}{rcl}
\ell\leq&\pi_{\inf}(p):=\liminf_{T\rightarrow\infty} \frac{1}{T}\sum_{t=1}^{T} {\pr[p\in \nu(S_t)]}&\\
&\pi_{\sup}(p):=\limsup_{T\rightarrow\infty} \frac{1}{T}\sum_{t=1}^{T} {\pr[{p\in \nu(S_t)}]}&\leq u
\end{array}\]
Intuitively, we measure the frequency with which $p$ is satisfied and the average frequency must remain in the constraining interval at all times (even when the limit does not exist), except for an initial ``heat-up'' phase.
It is worth noting that the policies we will synthesize always have even the limit defined. 
If an atomic proposition $p_s$ is only satisfied in state $s$, then the constraint $(s,x,x)$ is equivalent to the usual constraint $\pi(s)=x$ for the steady-state distribution $\pi$.

Further, $\mc$ \emph{$\delta$-satisfies} $\sss$ for some $\delta\geq0$, written $M\models_\delta\sss$, if $\mc\models\{(p,\ell-\delta,u+\delta)\mid (p,\ell,u)\in\sss\}$.
Intuitively, $\delta$-satisfaction allows for $\delta$ imprecision in satisfying the constraints.
Such a relaxation is often used, for instance, to obtain ``$\delta$-decidability'' of otherwise undecidable problems, e.g. on systems with complex continuous dynamics \cite{DBLP:conf/lics/GaoAC12,DBLP:journals/tac/RunggerT17}.
In our case, we show it allows for a polynomial-time solution even when restricting to finite-memory policies.



\para{Long-run average reward}
Let $\reward: A \to \Qset$ be a \emph{reward function}.  
Recall that $A_t$ is a random variable returning the action played at time $t$.
Similarly to the steady-state distribution, the random variable given by the limit-average function $\lrLim{\reward}=\lim_{T\rightarrow\infty} \frac{1}{T}\sum_{t=1}^{T} {\reward(A_t)}$ 
may be undefined for some runs, so we consider maximizing the respective point-wise limit inferior:
\[
\lrIf{\reward} = \liminf_{T\rightarrow\infty} 
\frac{1}{T}\sum_{t=1}^{T} {\reward(A_t)}
\]
Although we could also maximize limit superior, it is less interesting technically, see~\cite{lmcs14}, and less relevant from the perspective of ensuring the required performance. 
Further, the respective minimizing problems can be solved by maximization with opposite rewards.

\section{Problem Statement and Examples}\label{ssec:problem}

This paper is concerned with the following tasks:\medskip

\noindent
\framebox[0.5\textwidth]{\parbox{0.48\textwidth}{\smallskip
		
		\textbf{Satisfiability:}
		Given an MDP $\mdp$ with the \emph{long-run specification} $\lrs=((\varphi,\ps),\sss,(\reward,\rs))$, where $\varphi$ is an LTL formula, $\ps\in[0,1]$ is a probability threshold, $\sss$ is a steady-state specification, $\reward$ is a reward function, $\rs\in\Qset$ is a reward threshold, 
		decide whether there is a~policy $\sigma$ such that 
		\begin{align}
		\bullet\qquad& \Pr\sigma{}{\models\varphi}\geq \ps \tag{LTL}\label{eq:spec-ltl}\\
		\bullet\qquad& \mdp^\sigma\models\sss \tag{SSS}\label{eq:spec-sss}\\
		\bullet\qquad& \Ex{\sigma}{}{\lrIf{\reward}}\geq \rs \hspace*{15mm}\tag{LRA}\label{eq:spec-rew} 
		\end{align}
		\textbf{Policy synthesis:}
		If satisfiable, construct a~policy satisfying the requirements, i.e.\ inducing a Markov chain satisfying them.\smallskip
		
		\textbf{$\delta$-satisfying finite-memory policy synthesis} (for any $\delta>0$)\textbf{:}
		If satisfiable, construct a~finite-memory policy $\sigma$ ensuring
		\begin{align}
		\bullet\qquad& \Pr\sigma{}{\models\varphi}\geq \ps \tag{LTL}\label{eq:spec-dltl}\\
		\bullet\qquad& \mdp^\sigma\models_\delta\sss \tag{$\delta$-SSS}\label{eq:spec-dsss}\\
		\bullet\qquad& \Ex{\sigma}{}{\lrIf{\reward}}\geq \rs-\delta\tag{$\delta$-LRA}\label{eq:spec-drew} 
		\end{align}
	}}\smallskip

We illustrate the problem, some of its intricacies, the effect of the bound on the memory and the consequent importance of $\delta$-satisfaction on the following examples. 

\begin{example}\label{ex:below}
	Consider the MDP in Fig.~\ref{fig:MECs-below} with the specification $\pi(s)=0.5=\pi(t)$.
	On the one hand, every memoryless policy yields either $\pi(s)=1$ or $\pi(t)=1$, the latter occurring iff $b$ is played in $s$ with non-zero probability.
	On the other hand, a history-dependent policy $\sigma$ defined by $\sigma(sa\cdots s)(a)=1$ and $\sigma(\lambda)(a)=0.5=\sigma(\lambda)(b)$ satisfies the specification.
	In terms of the automata representation, the policy is 2-memory, remembering whether a step has been already taken, see Fig.~\ref{fig:MECs-below-auto} in Appendix~\ref{app:strat}.\qee
	
	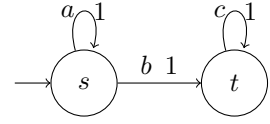
\begin{figure}[ht]
		\centering\spacefu
		\begin{tikzpicture}
		\node[state,initial,initial text=](s) at(0,0) {$s$};
		\node[state](t) at(2,0) {$t$};
		\path[->] 
		(s) edge[loop above] node[left]{$a$}node[right]{$1$}()
		(s) edge node[above]{$b\ \ 1$}(t)
		(t) edge[loop above] node[left]{$c$}node[right]{$1$}()
		;
		\end{tikzpicture}
		\caption{A (multichain) MDP where memory is required for satisfying a steady-state specification (SSS)}\label{fig:MECs-below}
	\end{figure}\spacefl
\end{example}
Consequently, memory may be necessary, in contrast to the claim of \cite{ijcai19} that memoryless policies are sufficient by \cite{akshay13}, which holds only for the setting with recurrent chains.
Moreover, the combination with LTL may require even unbounded memory:

\begin{example}
	Consider the MDP in Fig.~\ref{fig:MEC-switch} with the LTL specification $(\G\F p_t,1)$ where $p_t$ holds in $t$ only, and steady-state constraint $(s,1,1)$.
	Every policy satisfying the specification must play action $b$ infinitely often.
	On the one hand, a history-dependent policy $\sigma$ defined by $\sigma(\fpat s)(b)=1/|\fpat|$, where $|\fpat|$ denotes the length of $\fpat$, satisfies the specification.
	Indeed, $t$ is still visited infinitely often almost surely, but with frequency decreasing below every positive bound.
	While $\sigma$ is Markovian (does not depend on the history, only on the time), it means it still uses unbounded memory.
	
	On the other hand, every policy with finite memory satisfying the LTL specification visits $t$ with positive frequency, in fact with at least $1/p^{|S|\cdot m}$ where $m$ is the size of the memory and $p$ the minimum probability occurring in the policy. 
	Hence no finite-memory policy can satisfy the whole specification.
	However, in order to $\delta$-satisfy it, a memoryless policy is sufficient, defined by $\sigma(s)(b)=1/\delta$. \qee
	
	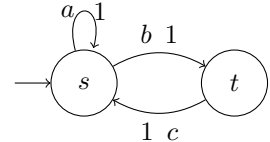
\begin{figure}[ht]
		\centering\spacefl
		\begin{tikzpicture}
		\node[state,initial,initial text=](s) at(0,0) {$s$};
		\node[state](t) at(2,0) {$t$};
		\path[->] 
		(s) edge[loop above] node[left]{$a$}node[right]{$1$}()
		(s) edge[bend left] node[above]{$b\ \ 1$}(t)
		(t) edge[bend left] node[below]{$1\ \ c$}(s)
		;
		\end{tikzpicture}\spacefl
		\caption{A (unichain) MDP where unbounded memory is required for satisfying SSS and LTL specifications, but finite memory is sufficient for $\delta$-satisfaction}
		\label{fig:MEC-switch}
	\end{figure}\spacefl
\end{example}

%
%

\section{LTL-Constrained Steady-State Policy Synthesis}

Fig.~\ref{fig:lp-old} depicts a set of linear constraints capturing (both transient and recurrent) behaviour of a policy as a ``flow'' \cite{lmcs17}, with variables capturing how much each action is used.
Together with maximization of a certain linear function it yields the standard linear programme for maximizing LRA reward in general (multichain) MDP \cite{puterman}.
Since then it has been used in various contexts, including multiple reachability objectives \cite{EKVY08} or multi-dimensional rewards \cite{lmcs14,lmcs17}.

\begin{figure}[t]
	\framebox[\columnwidth]{\parbox{0.96\columnwidth}{\smallskip
	\noindent{\bf Input:} MDP $(S,A,\mathit{Act},\delta,\inits,\nu)$	\medskip	
			
	Beside requiring all variables $y_a,y_s,x_a$ for $a\in A,s\in S$ be non-negative, the \emph{policy-flow} constraints are the following: \bigskip
	
	\begin{enumerate}
		\item transient flow: for $s\in S$
		$$\mathbf 1_{\inits}(s)+\sum_{a\in A}y_a\cdot \delta(a)(s)=\sum_{a\in \act{s}} y_a+y_s$$
		\item almost-sure switching to recurrent behaviour:
		$$\sum_{s\in C\in \mec}y_s=1$$
		\item probability of switching in a MEC is the frequency of using its actions: for $C\in \mec$
		$$\sum_{s\in C}y_s=\sum_{a\in C}x_a$$
		\item recurrent flow: for $s\in S$
		$$\sum_{a\in A}x_a\cdot \delta(a)(s)=\sum_{a\in \act{s}} x_a$$
	\end{enumerate}
}}
	\caption{Linear program for policy synthesis in general (multichain) MDP }
	\label{fig:lp-old}
\end{figure}

Intuitively, $x_a$ is the expected frequency of using $a$ on the long run; Equation 4 thus expresses the recurrent flow in MECs and $\pi(s):=\sum_{a\in\act{s}}x_a$ is the steady-state distribution for state $s$. 
However, before we can play according to $x$-variables, we have to reach MECs and switch from the transient behaviour to this recurrent behaviour. 
Equation~1 expresses the transient flow before switching. 
Variables $y_a$ are the expected number of using $a$ until we switch to the recurrent behaviour in MECs and $y_s$ is the probability of this switch upon reaching $s$.
To relate $y$- and $x$-variables, Equation 3 states that the probability to switch and remain within a given MEC is the same whether viewed from the transient or recurrent flow perspective.\footnote{Actually, one could eliminate variables $y_s$ and use directly $x_a$ in Equation~1 and leave out Equation 3 completely, in the spirit of \cite{puterman}.
	However, the form with explicit $y_s$ is closer to the more recent related work, more intuitive, and more convenient for correctness proofs.}  
Finally, Equation 2 states that switching happens almost surely.\footnote{Note that summing Equation 1 over all $s\in S$ yields $\sum_{s\in S}y_s=1$. 
Since $y_s$ can be shown to equal $0$ for state $s$ not in MEC, Equation 2 is again redundant but more convenient.}

\begin{lemma}[\cite{lmcs17}]
	Existence of a solution to the equations in Fig.~\ref{fig:lp-old} is equivalent to existence of a policy inducing the steady-state distribution $\pi(s)=\sum_{a\in\act{s}}x_a$ for all $s\in S$.\qed
\end{lemma}

%
%

\begin{figure}	
	\framebox[\columnwidth]{\parbox{0.96\columnwidth}{\smallskip
			\noindent{\bf Input:} MDP $(S,A,\mathit{Act},\delta,\inits,\nu)$,	
			long-run specification $\lrs=((\varphi,\ps),\sss,(\reward,\rs))$, 
			$\amec$ due to the assumption that the MDP is a product of another MDP with $\auto$
			\medskip
			
			The \emph{specification constraints} are the following:
			\bigskip
			
			\begin{enumerate}\setcounter{enumi}4
				\item\label{eq:ltl} LTL specification: 
				$$\sum_{a\in A\setminus\bigcup\amec} x_a\leq1-\ps$$
				
				\item\label{eq:ss} steady-state specification: for every $(p,\ell,u)\in\sss$
				$$\ell\leq\sum_{\substack{p\in\nu(s)\\a\in \act{s}}}x_a\leq u$$
							
				\item\label{eq:rew} LRA rewards:
				$$\sum_{a\in A}x_{a}\cdot\reward(a)\geq \rs$$
			\end{enumerate}
		}}
	\caption{Linear program for the long-run specification on the product MDP (together with the policy-flow constraints of the LP in Fig.~\ref{fig:lp-old})}
	\label{fig:lp-new}
\end{figure}
		
\begin{theorem}[Linear program for LTL-constrained steady-state policy synthesis on general MDP]\label{thm}		

	For an MDP $\mdp$ and a long-run specification $\lrs$, let $L$ be the system consisting of linear policy-flow constraints of Fig.~\ref{fig:lp-old} and specification constraints of Fig.~\ref{fig:lp-new} both on input $\mdp\times\auto$ where $\varphi$ is the LTL formula of $\lrs$. Then:
	\begin{enumerate}
		\item Every policy satisfying $\lrs$ induces a solution to $L$.
		\item Every solution to $L$ effectively induces a (possibly unbounded-memory) policy satisfying $\lrs$ and, moreover for every $\delta>0$, also a finite-memory policy $\delta$-satisfying $\lrs$ .
	\end{enumerate}		
\end{theorem}

\begin{proof}[Proof sketch]
	The full proof can be found in 
	\ifdefined\arxiv
	Appendix~\ref{app:proof}. 
	\else
	\cite[Appendix B]{techrep}. 
	\fi
	It heavily relies on results about the linear programming solution for the multi-dimensional long-run average reward \cite{lmcs17} and also on the LTL model checking algorithm on MDP via LDBA~\cite{cav16}. 
	Essentially, each steady-state constraint can be seen as a LRA constraint on a new reward that is 1 whenever $p$ holds and 0 otherwise.
	Besides, $\delta$-satisfaction of the LTL constraint with finite memory is equivalent to positive reward collected on accepting states.
	
	The constructed policy first reaches the desired MECs where it ``switches'' to staying in the MEC and playing a memoryless policy whose frequencies of using each action $a$ are close to $x_a$, and hence we get a 2-memory policy; alternatively, the switch might be followed by a policy achieving exactly $x_a$ but possibly requiring unbounded memory.
	The policy's distribution over actions then depends on the length of the current history. 
\end{proof}

\begin{corollary}
	Given an MDP $\mdp$ with a long-run specification $\lrs$ (possibly with a trivial reward threshold, e.g. the minimum reward), a policy maximizing reward among policies satisfying $\lrs$ can be effectively synthesized.
\end{corollary}
\begin{proof}
	First, we construct the product $\mdp\times\auto$, where $\varphi$ is the LTL formula of $\lrs$.
	Second, we construct the linear programme with the objective function $$\max \sum_{a\in \act{s}}x_a\cdot\reward(a)$$
	and the constraints 1.--7., or in the case of trivial reward threshold only 1.--6.
	Thirdly, a solution to the programme induces the satisfying policy (as described in detail in the proof of Theorem~\ref{thm}.
\end{proof}

\section{Complexity}
Observe that the \emph{number of variables} is linear in the size of the product and  the \emph{size} of the LP is quadratic, hence the size is polynomial in the size of $\mdp$ and $\auto$.
Since linear programming is solvable in polynomial time \cite{LP}, our problems can also be solved in time polynomial in $\mdp$ and $\auto$.

Consequently, the satisfiability problem can be solved in double exponential time due to the translation of LTL to LDBA with this complexity.
In fact, it is 2-EXPTIME-complete, inheriting the hardness from its special case of LTL model checking of MDP.
While this sounds pessimistic at first glance, 
the worst-case complexity is rarely any issue in practice whenever it stems only from the translation as here.
Indeed, the sizes of the LDBA produced for practically occurring formulae are typically very small, see \cite{cav16,mochiba,rabinizer,DBLP:journals/jacm/EsparzaKS20} for examples and more details.
Moreover, in the frequent case of qualitative (almost sure) LTL satisfaction, Constraint~\ref{eq:ltl} takes the form: for every $a\in A\setminus\bigcup\amec:x_a=0$, effectively eliminating many variables from the LP.

Further, for long-run specifications with $\omega$-regular properties (instead of LTL), the problem can be solved in polynomial time if the input specification is provided directly as an LDBA.
If instead a fully non-deterministic BA is used, single exponential time is required due to the semi-determinization to LDBA \cite{CY95}.

In any case, the algorithm is polynomial with respect to the MDP and there are no extra variables for the steady-state or reward specifications.
Hence, if the LTL property is trivial (omitted), there is almost no overhead compared the LP of Fig.~\ref{fig:lp-old} (or the LP for rewards which contains one constraint (\ref{eq:rew}.) more).
In particular, for point specifications with constraints $(p,x,x)$ as in \cite{akshay13}, Constraint~\ref{eq:ss} takes the form $\sum_a x_a=x$, which can be substituted into the system, even decreasing the number of its variables.

The final remark is concerned with the type of the automaton used.
While we used LDBA, one could use other, deterministic automata such as the traditionally used Rabin or parity \cite{BK08}.
However, there are two drawbacks of doing so.
Firstly, while the translation from LTL is for all the cases double exponential in the worst case, practically it is often worse for Rabin than for LDBA and yet significantly worse for parity, see e.g.~\cite{rabinizer}.
Secondly, the acceptance condition is more complex.
Instead of the B\"uchi condition with infinitely many visits of a set of states, for Rabin there is disjunction of several options, each a conjunction of the B\"uchi condition and prohibiting some other states.
A reduction to LRA would require several copies of these parts in the product, each with the prohibited parts removed, resulting in a larger system.
For parity, one copy would be sufficient together with a pre-processing by the attractor construction, yet, as already mentioned, the automaton is typically significantly larger than the LDBA, see \cite{tacas17-ldba}.

\section{Discussion}

We have seen we can synthesize satisfying policies if they exist.
The generality of the framework can be documented by the ease of extending our solution to various modifications of the problem.
For instance, the general policies suffer from several deficiencies, which we can easily address here:
\begin{itemize}
	\item \emph{Unbounded memory} may not be realistic for implementation.
	As discussed, we can solve the $\delta$-satisfaction problem in polynomial time for steady-state specifications (or also for general long-run specification provided the $\auto$ is given on input), yielding finite-memory policies, where the imprecision $\delta$ can be arbitrarily small.
	\item The unbounded-memory solution may lead to policies visiting the accepting states \emph{less and less frequently} over time.
	Avoiding this is easy through an additional constraint
	$$\sum_{a\in \acc\cap C\in\amec}x_a  \geq \frac1f$$
	ensuring the frequency does not decrease below once per $f$ steps on average.
	Alternatively, if we want to enforce the frequency bound for almost all satisfying runs, then we can instead use the frequency bound in each MEC: 
	$$\sum_{a\in \acc\cap C}x_a  \geq \frac1f\cdot \sum_{a\in A\cap C}x_a$$	
\end{itemize}

Further modifications and their solutions include:
\begin{itemize}
	\item Similarly to the frequency bounds for visiting accepting states, we may require bounds on \emph{frequency of satisfying recurrent subgoals}, i.e.\ subformulae of $\varphi$.
	Indeed, the translation from LTL to LDBA \cite{cav16} labels the states with the progress of satisfying subformulae.
	For each subformula, there are its ``breakpoint'' states where the subformula is satisfied, and whose $x$ variables we can constrain as above.
	
	\item LRA rewards can be \emph{multi-dimensional}, i.e., take the vector form $\vec{\reward} : A \to \Qset^n$.
	Then the single Constraint~\ref{eq:rew} appears once for each dimension, with negligible additional computational requirements.
	Moreover, we can optimize any weighted combination of the LRA rewards or consider the trade-offs among them.
	To this end, we can $\varepsilon$-approximate the Pareto frontier capturing all pointwise optimal combinations, in time polynomial in the size of the product and in $1/\varepsilon$ \cite{PY00,lmcs17}.
	
	\item We can \emph{optimize the satisfaction probability of the LTL} formula instead of the reward 
	or consider the trade-offs.
	To this end, it is important that our solution uses a single linear programme.
	A possible approach of analyzing each MEC separately and then merging the fixed solutions would suffer from the mutual inflexibility and could not efficiently address these trade-offs.
	
	\item In contrast, if only \emph{unichain}\footnote{A Markov chain is a unichain if it only has one bottom strongly connected component (one recurrent class) and all other states are transient.} 
	solutions should be considered, see \cite{ijcai19}, we can check each accepting MEC separately (through our LP) whether it also satisfies the steady-state specification and whether it can be reached almost surely.

\end{itemize}

As future work, one may consider combinations with non-linear properties, such as those expressible in branching-time logics, or with finite-horizon or discounted rewards.

\section*{Acknowledgments}
This work has been partially supported by the German Research Foundation (DFG) projects 427755713 (KR 4890/3-1) \emph{Group-By Objectives in Probabilistic Verification (GOPro)} and 317422601 (KR 4890/1-1) \emph{Verified Model Checkers (VMC)}.

\ifdefined\arxiv
\bibliographystyle{named}
\bibliography{ijcai21}
\else
\fi

\appendix

\section*{Appendix}

\section{Finite-Memory Policies}\label{app:strat}

In order to work with finite-memory policies,
we also use a slightly
different (though equivalent---see~\cite[Section 6]{lmcs14})
definition of policies, which is more convenient in some settings. 
Let $\mem$ be a countable set of \emph{memory elements}. 
A \emph{policy} is a triple
$\sigma = (\sigma_u,\sigma_n,\alpha)$, where 
$\sigma_u: A\times S \times \mem \to \dist(\mem)$ and 
$\sigma_n: S \times \mem \to \dist(A)$ are \emph{memory update}
and \emph{next move} functions, respectively, and $\alpha$ is
the initial distribution on memory elements. We require that, for 
all $(s,m) \in S \times \mem$, the distribution $\sigma_n(s,m)$ assigns a
positive value only to actions enabled at~$s$, i.e.\ $\sigma_n(s,m)\in\dist(\act{s})$. 

A \emph{play} of $\mdp$ determined by 
a policy $\sigma$ is a Markov chain 
$\mdp^\sigma$ where the set of locations is $S \times \mem \times A$,
the initial distribution $\mu$ is zero except for 
$\mu(\inits,m,a) = \alpha(m) \cdot \sigma_n(\inits,m)(a)$, 
and
\ifdefined\lncs
\[ 
P(s,m,a)(s',m',a')\ =\ 
\delta(a)(s')  \cdot \sigma_u(a,s',m)(m') \cdot \sigma_n(s',m')(a')\,.
\]
\else
\begin{multline*} 
P(s,m,a)(s',m',a')\ =\ \\ 
\delta(a)(s')  \cdot \sigma_u(a,s',m)(m') \cdot \sigma_n(s',m')(a')\,. 
\end{multline*}
\fi
Hence, $\mdp^\sigma$ starts in a location chosen randomly according
to $\alpha$ and $\sigma_n$. In a current location $(s,m,a)$, 
the next action to be performed is $a$, hence the probability of entering
$s'$ is $\delta(a)(s')$. The probability of updating the memory to $m'$
is $\sigma_u(a,s',m)(m')$, and the probability of selecting $a'$ as 
the next action is $\sigma_n(s',m')(a')$. Note that these choices
are independent, and thus we obtain the product above.

In general, a policy may use infinite memory $\mem$, and both 
$\sigma_u$ and $\sigma_n$ may randomize, the former called \emph{stochastic-update}, the latter \emph{randomization} in the narrower sense. 

We can now classify the policies according to the size of memory
they use. Important subclasses are 
\emph{memoryless} policies, in which $\mem$ is a singleton,
\emph{$n$-memory} policies, in which $\mem$ has exactly $n$~elements, and
\emph{finite-memory} policies, in which $\mem$ is finite.
Other policies are \emph{unbounded-memory} or \emph{infinite-memory} policies.

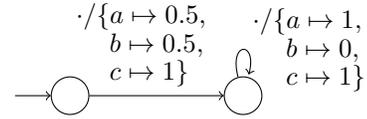
\begin{figure}[t]
	\centering
	\begin{tikzpicture}
	\node[state,initial,initial text=,minimum size=5mm](s) at(0,0) {};
	\node[state,minimum size=5mm](t) at(2.3,0) {};
	\path[->] 
	(s) edge node[above,pos=0.6,text width=25mm]{$\cdot/\{a\mapsto0.5,\newline \phantom{\cdot/\{}b\mapsto0.5,\newline\phantom{\cdot/\{} c\mapsto1\}$}(t)
	(t) edge[loop above] node[right,pos=0.5,text width=20mm]{$\cdot/\{a\mapsto1,\newline \phantom{\cdot/\{}b\mapsto0,\newline\phantom{\cdot/\{} c\mapsto1\}$}()
	;
	\end{tikzpicture}
	\caption{An automaton representation of the 2-memory strategy considered in Example~\ref{ex:below} depicted in Fig.~\ref{fig:MECs-below}}\label{fig:MECs-below-auto}
\end{figure}


%

\ifdefined\arxiv

\section{Proof of Correctness of Theorem~\ref{thm}}\label{app:proof}

\begin{proof}
	(1.
	):
	Let $\sigma$ be a satisfying policy on $\mdp$.
	By \cite[Theorem~3]{cav16}, there is a policy $\sigma'$ on $\mdp\times\auto$ that is (i) identical to $\sigma$ on $\mdp$ (not using the history on the $\auto$ component, except for doing the jump on $\auto$ upon arrival to the bottom strongly connected component of $\mdp^\sigma$), (ii) making runs of measure at least $\theta$ reach $\amec$ of $\mdp\times\auto$, remain there and visit its set of accepting states infinitely often.

	To formalize the frequency of using an action $a$ on a run, recall $\kda(a)=1$ and $\kda(b)=0$ for $a\neq b\in A$. 
	Then $\lrIf{\kda}$ defines a random variable for each $a\in A$ (where $A$ are actions of the product).
	By \cite[Section~4]{lmcs17}, $\sigma'$ induces a solution to Equations 1.--4.\ and Constraint \ref{eq:rew}.\ such that (iii) $\sum_{a\in A}x_a=1$, and for each $a$, $x_a\in[\lrIf{\kda},\lrSf{\kda}]$.
	For $(p,\ell,u)\in\sss$, we sum this up over all actions leaving states labelled with $p$, yielding
	$$\sum_{\substack{p\in\nu(s)\\a\in \act{s}}}x_a\in[\pi_{\inf}'(p),\pi_{\sup}'(p)]=[\pi_{\inf}(p),\pi_{\sup}(p)]$$
	with the latter equality by (i).
	Since the policy satisfies $\sss$, the latter interval is a subset of $[\ell,u]$, implying Constraint \ref{eq:ss}.
	
	Finally, Constraint~\ref{eq:ltl} is also satisfied since runs of measure at least $\theta$ have suffixes in $\amec$ by (ii).
	Hence the long-run frequency of actions inside $\amec$ is $\sum_{a\in \bigcup\amec} x_a\geq\ps$.
	By (iii), Constraint~\ref{eq:ltl} follows.
	\medskip
	

	(2.
	):
	We transform the solution to the constraints into the satisfying policies. 
	We shall work with policies where the frequencies of using actions are well defined and do not oscillate.
	Hence we define $\vfreq_a:=\lrLim{\kda}\equiv\lim_{T\rightarrow\infty} 
	\frac{1}{T}\sum_{t=1}^{T} {\kda(A_t)}$ whenever the limit exists.
	
	Let us now focus on the case of general policies (with no bounds on memory).
	We start with the recurrent part.
	By \cite[Corollary 5.5]{lmcs17}, the solution to Eq.~4 yields for each MEC $C$ of $\mdp\times\auto$ with non-zero $\sum_{a\in A\cap C}x_a$ a~Markovian policy $\sigma_C^0$  with a.s.\  $\freq_a=x_a/\sum_{a\in A\cap C}x_a$ and visiting all states of $C$.
	
	By \cite[Corollary 5.8]{lmcs17} and Equations~1--3, $\sigma_C^0$'s can be ``linked'' together into  $\sigma^0$ satisfying $\freq_a=x_a$.
	Intuitively, a memoryless policy given by $y_a$'s is played until the moment we should remain in the current MEC $C$ (with the respective probability $\sum_{s\in S\cap C}y_s$), at which point $\sigma^0$ switches to playing the respective $\sigma_C^0$.
	Consequently, for $(p,\ell,u)\in\sss$,
	$$\freq_p=\sum_{\substack{p\in\nu(s)\\a\in \act{s}}} \freq_a = \sum_{\substack{p\in\nu(s)\\a\in \act{s}}} x_a\in[\ell,u]$$
	with the latter by Constraint~\ref{eq:ss} and hence $\sss$ is satisfied.
	
	By the well defined frequencies, we get 
	$$\lrIf{\reward}=\sum_{a\in\bar A}\freq_a\cdot\reward(a) = \sum_{a\in\bar A}x_a\cdot\reward(a)\geq r$$
	with the latter by Constraint~\ref{eq:rew}, hence the reward specification is also satisfied.
	
	By Constraint~\ref{eq:ltl} and $\freq_a= x_a$,\footnote{Recall that in general $\freq_a=0$ does not mean that $a$ cannot be played infinitely often, even almost surely, but only that in such a case the frequency steadily decreases.} $\theta$-portion of the time is spent playing inside $\amec$.
	Hence at least $\theta$-portion of runs play in $\amec$ infinitely often.
	Since there are only finitely many MECs a run can go through, each of these runs must remain in a single accepting MEC eventually.
	So the probability of remaining in $\amec$ altogether is at least $\theta$.
	Moreover, by \cite[Remark 5.6]{lmcs17}, all states of the MEC $C$ are visited by $\sigma^0_C$ infinitely often (on almost all runs remaining in $C$), in particular also the accepting states, hence $\varphi$ is satisfied with probability at least $\theta$.
	\medskip

	Let us now consider the case of bounded-memory policies now.
	Instead of $\sigma_C^0$'s, we use \cite[Lemma 5.3]{lmcs17} to produce memoryless policies $\sigma_C^\varepsilon$ with a.s. $\freq_a$ equal to $x_a/\sum_{a\in A\cap C}x_a$ up to $\varepsilon$ and visiting all states of $C$ infinitely often.
	Similarly, 	again by \cite[Corollary 5.8]{lmcs17}, upon reaching the desired MECs (using a memoryless policy) one can switch to $\sigma_C^\varepsilon$ (instead of $\sigma^0_C$), yielding a policy $\sigma^\varepsilon$ with $\freq_a$ equal to $x_a$ up to the error of $\varepsilon$ (we choose an appropriate $\varepsilon$ below).
	
	Consequently, for $(p,\ell,u)\in\sss$, we have that $\freq_p=\sum_{\substack{p\in\nu(s)\\a\in \act{s}}} \freq_a$ is $\sum_{\substack{p\in\nu(s)\\a\in \act{s}}} x_a$ up to the error $\delta=|A|\cdot\varepsilon\leq |A_{\mdp}|\cdot|Q|\cdot\varepsilon$, where $|A_\mdp|$ is the number of actions of the original MDP (not the product).
	Hence, for the required $\delta$-satisfiability of $\sss$, it is sufficient to choose $\varepsilon=\delta/(|Q|\cdot|A_\mdp|)$ (or smaller).
	
	By the well defined frequencies,
	$$\lrIf{\reward}=\sum_{a\in\bar A}\freq_a\cdot\reward(a) \geq \sum_{a\in \bar A}(x_a-\varepsilon)\cdot\reward(a)=:(*)$$
	and by Constraint~\ref{eq:rew}
	$$(*)\geq r-\varepsilon|A|\cdot||\reward||_\infty \geq r-\varepsilon\cdot|Q|\cdot|A_\mdp|\cdot \rmax$$
	where $\rmax$ is the maximum reward.
	Hence the reward and other specifications are $\delta$-satisfied if we choose $$\varepsilon=\delta/(|Q|\cdot|A_\mdp|\cdot\max\{1,\rmax\})$$ (or smaller).
	Indeed, for the LTL specification the argument above applies: although $\freq_a$ may differ from $x_a$, the overall portion of the time spent playing inside $\amec$ is still at least $\theta$ (with no error induced).
	
	Note that due to the single switching moment and due to $\sigma_C^\varepsilon$ being memoryless, $\sigma^\varepsilon$ is a finite-memory policy (actually a 2-memory stochastic-update policy in the sense of Appendix~\ref{app:strat}).
\end{proof}

\fi

\ifdefined\arxiv
\else
\bibliographystyle{named}
\bibliography{ijcai21}
\fi

%
%
%
\end{document}